%% file: KMeans-arXiv.tex
\documentclass{article}

\usepackage[preprint]{nips_2018}

\usepackage[utf8]{inputenc} 
\usepackage[T1]{fontenc}    
\usepackage{hyperref}       
\usepackage{url}            
\usepackage{booktabs}       
\usepackage{amsfonts}       
\usepackage{nicefrac}       
\usepackage{microtype}      
\usepackage{amsmath}
\usepackage{multirow}
\usepackage[ruled,linesnumbered]{algorithm2e}

\usepackage{amssymb}
\usepackage{bm}
\usepackage{textcomp}
\usepackage{bigstrut}
\usepackage{array}
\usepackage{gensymb}
\usepackage{pifont}
\usepackage{graphicx}
\usepackage{psfrag}
\usepackage[mathscr]{euscript}
\usepackage{enumitem}
\usepackage{rotating}
\usepackage{enumitem}
\usepackage{makecell}

\DeclareMathOperator*{\argmin}{arg\,min}

\newcommand{\assignmentstep}{\emph{Assignment-Step}}
\newcommand{\updatestep}{\emph{Update-Step}}

\title{Fast K-Means Clustering with Anderson Acceleration}

\author{
  Juyong Zhang \quad\qquad Yuxin Yao \quad\qquad Yue Peng \quad\qquad Hao Yu\\
  University of Science and Technology of China\\
  \texttt{juyong@ustc.edu.cn~~yaoyuxin1@126.com~~\{echoyue,yh8971\}@mail.ustc.edu.cn}\\
  \And
  Bailin Deng\thanks{Corresponding author.}\\
  Cardiff University\\
  \texttt{DengB3@cardiff.ac.uk}
}

\begin{document}

\maketitle

\begin{abstract}
We propose a novel method to accelerate Lloyd's algorithm for K-Means clustering. Unlike previous acceleration approaches that reduce computational cost per iterations or improve initialization, our approach is focused on reducing the number of iterations required for convergence. This is achieved by treating the assignment step and the update step of Lloyd's algorithm as a fixed-point iteration, and applying \emph{Anderson acceleration}, a well-established technique for accelerating fixed-point solvers.
Classical Anderson acceleration utilizes $m$ previous iterates to find an accelerated iterate, and its performance on K-Means clustering can be sensitive to choice of $m$ and the distribution of samples. We propose a new strategy to dynamically adjust the value of $m$, which achieves robust and consistent speedups across different problem instances. Our method complements existing acceleration techniques, and can be combined with them to achieve state-of-the-art performance. We perform extensive experiments to evaluate the performance of the proposed method, where it outperforms other algorithms in 106 out of 120 test cases, and the mean decrease ratio of computational time is more than 33\%.
\end{abstract}

\input{introduction}
\input{problem}

\input{results}

\input{conclusion}

\bibliographystyle{natbib}
\bibliography{kmeans}

\end{document}

%% file: introduction.tex
\section{Introduction}

K-Means clustering is a fundamental method with various applications such as data compression, classification, and density estimation. Given a set of $N$ samples $\mathbf{x}_{1}, \ldots, \mathbf{x}_N \in \mathbb{R}^{d}$, and a positive number $K \leq N$, K-Means clustering partitions the samples into $K$ clusters, by minimizing the total squared distances from the samples to the centroids $\mathbf{c}_1, \ldots, \mathbf{c}_K \in \mathbb{R}^d$ of their respective clusters, resulting in an optimization problem 
\begin{equation}
\min_{\mathbf{C}} \; E(\mathbf{C}) = \sum_{i=1}^{N} \|\mathbf{x}_i - \mathbf{c}_{\rho_i}\|^2,
\label{globalEnergy}
\end{equation}
where $\mathbf{C} = \left[\mathbf{c}_1, \ldots, \mathbf{c}_K\right]$, and $\mathbf{c}_{\rho_i}$ is the centroid closest to the sample $\mathbf{x}_i$.
This is a non-convex optimization problem, and it is NP-hard to find the global minimum~\citep{Aloise2009}. On the other hand, there exist effective algorithms to find a local minimum, the most popular being \emph{Lloyd's algorithm}, often simply referred to as the K-Means algorithm.

The classical Lloyd's algorithm is equivalent to minimizing the target energy with both the cluster assignment and the centroid positions as variables:
\begin{equation}
	\min_{\{\rho_i\}, \{\mathbf{c}_j\}} \sum_{i=1}^{N}\|\mathbf{x}_i - \mathbf{c}_{\rho_i}\|^2,
	\label{eq:LloydsMinimization}
\end{equation}
where $\rho_i$ ($i = 1, \ldots, N$) is the index of the cluster that $\mathbf{x}_i$ is assigned to.  
Lloyd's algorithm performs the minimization in a two-step iterative process: 
\begin{itemize}[leftmargin=*]
	\item In the \emph{assignment} step, each sample $\mathbf{x}_i$ is assigned to the cluster whose centroid is the closest to $\mathbf{x}_i$:
	\begin{equation}
		\rho_i^{t+1} = \argmin_{j\in \{1,\ldots,K\}}\|\mathbf{x}_i - \mathbf{c}_j^{t}\|.
		\label{eq:AssignmentStep}
	\end{equation}
	\item In the \emph{update} step, each centroid $\mathbf{c}_j$ is updated to the mean of all samples assigned to cluster $j$:
	\begin{equation}
		\mathbf{c}_j^{t+1} = \frac{1}{N_j^{t+1}} \sum_{\{i \mid \rho_i^{t+1} = j\}} \mathbf{x}_i,
		\label{eq:UpdateStep}
	\end{equation}
	where $N_j^{t+1}$ is the number of assigned samples for cluster $j$.
\end{itemize}
This iterative process can be considered as alternating minimization of the target function with respective to the assignment variables and the centroid variables, respectively. As a result, Lloyd's algorithm monotonically decreases the target function and guarantees convergence. 
In practice, however, Lloyd's algorithm can suffer from slow convergence rate and require a large number of iterations. Different acceleration techniques have been proposed to improve its performance:
\begin{itemize}[leftmargin=*]
\item In each iteration of Lloyd's algorithm, the assignment step is the main computational bottleneck: a na\"{i}ve implementation would require $O(NK)$ distance computations to determine the nearest centroids for all samples, which can be a significant cost especially for a large number of samples or clusters. Therefore, many existing works focus on reducing the computational cost per iteration, e.g., by maintaining distance bounds between samples and centroids to avoid unnecessary calculations~\citep{elkan2003using,hamerly2010making,hamerly2015accelerating,ding2015yinyang,newling2016fast}.
\item As the target function of K-Means clustering is non-convex, different initial centroid positions might lead to different clustering results and different number of iterations used in Lloyd's algorithm. Various algorithms aim at improving the initialization to achieve better clustering and reduced computational time~\citep{NgH94,bradley1998refining,arthur2007k,bachem2016fast,newling2017k}.
\end{itemize} 

Despite the improved performance provided by these methods, they do not alter the convergence rate of Lloyd's algorithm and may still suffer from slow convergence. In this paper, we propose a novel approach that accelerates Lloyd's algorithm from a different perspective. Our approach aims at modifying the iterates from Lloyd's algorithm to improve the convergence rate and reduce the number of required iterations. Our contribution is two-fold:
\begin{itemize}[leftmargin=*]
	\item First, we treat the sequence of centroid positions generated by Lloyd's algorithm as the results of a fixed-point iteration solver. This enables us to improve their convergence rate using \emph{Anderson acceleration}~\citep{Anderson1965,Walker2011}, a well-established technique for accelerating fixed-point solvers~\citep{Fang2009,Toth2015,Higham2016,Toth2017}. We adopt the modification proposed by~\cite{Peng18AA} to reduce the computational overhead and improve stability. The resulting solver greatly improves the performance of Lloyd's algorithm in many instances. 
	\item Moreover, we show that the performance of classical Anderson acceleration, which relies on the previous $m$ iterates to determine an accelerated iterate, can be further improved by dynamically adjusting the value of $m$ according to the decrease of target energy. Using this strategy, the accelerated solver achieves consistent acceleration across all problem instances and initializations in our experiments.
\end{itemize}

Our approach can not only be used as a standalone technique to accelerate Lloyd's algorithm, but also be used in conjugation with existing approaches that reduce computational cost per iteration and improve initialization to deliver state-of-the-art performance for K-Means clustering. The effectiveness of our approach is tested on a large variety of datasets.

%% file: problem.tex
\section{Algorithm}
Before presenting our acceleration technique, let us first analyze the classical Lloyd's algorithm to see why it can suffer from slow convergence. The main strength of Lloyd's algorithm is the monotonic decrease of target energy and the resulting guarantee of convergence. This property stems from the fact that each iteration of Lloyd's algorithm computes the new centroids $\{\mathbf{c}_j^{t+1}\}$ by minimizing a convex surrogate (i.e., upper bound function) of the target energy $E$ in~\eqref{globalEnergy} that is constructed from the current centroid positions $\{\mathbf{c}_j^{t}\}$. Specifically, using the correspondence between samples and centroids determined in the assignment step~\eqref{eq:AssignmentStep}, we can derive a surrogate of $E$
\begin{equation}
	\overline{E}^{t} (\mathbf{C}) = \sum_{i=1}^N \|\mathbf{x}_i - \mathbf{c}_{\rho_i^{t+1}}\|^2 = \sum_{j=1}^K \sum_{i \in \mathcal{S}_{j}^{t}} \|\mathbf{x}_i - \mathbf{c}_j\|^2, 
	\label{eq:Surrogate}
\end{equation}
where $\mathcal{S}_{j}^{t}$ is the index set of samples assigned to centroid $\mathbf{c}_j^t$. In particular, it can be shown that
\[
	E(\mathbf{C}) \leq \overline{E}^t (\mathbf{C})~~\forall\mathbf{C},\quad E(\mathbf{C}^{t}) = \overline{E}^t(\mathbf{C}^{t}).
\] 
The minimization of this surrogate function is exactly the update step~\eqref{eq:UpdateStep}. The surrogate property then guarantees the decrease of target energy $E$:
\[
	E(\mathbf{C}^{t+1}) \leq \overline{E}^t(\mathbf{C}^{t+1}) \leq \overline{E}^t(\mathbf{C}^{t}) = E(\mathbf{C}^{t}).
\] 
From this perspective, Lloyd's algorithm is an instance of the majorization-minimization algorithm~\citep{Lange2016}, which performs energy minimization by iteratively constructing and minimizing surrogate functions.

On the other hand, the surrogate function can be both a blessing and a curse. It is only guaranteed to be accurate within a small neighborhood of the current iterate $\mathbf{C}^{t}$, and can deviate significantly from the true energy when being far away from $\mathbf{C}^{t}$. Such deviation is incurred when changed centroid positions alter the assignment of samples to centroids, which occurs frequently unless the samples are highly separated into different clusters. In other words, there may be a rapid loss of accuracy for the surrogate function when moving away from $\mathbf{C}^{t}$, which can hinder the energy decrease in the update step and leads to slow convergence.

Indeed, it has been pointed out by~\cite{BottouB94} that locally Lloyd's algorithm is equivalent to Newton's algorithm for minimizing the energy $E$. Since infinitesimal changes of centroid positions do not alter the sample-centroid assignment, the surrogate function~\eqref{eq:Surrogate} is the second-order Taylor expansion of the target energy, and its minimization is equivalent to a Newton step. If the new centroid positions $\mathbf{C}^{t+1}$ result in the same sample assignment as $\mathbf{C}^{t}$, then the Newton algorithm converges in one step and $\mathbf{C}^{t+1}$ is a local minimum. However, in many instances a large step will result in changes of sample assignment and invalidates the Taylor expansion, which slows down the convergence. 

From the discussions above, we can see that an important reason for slow convergence of Lloyd's algorithm is that the centroid update only considers the local landscape of the target energy. Therefore, a natural idea to improve convergence is to incorporate more global information, which is the essence of the technique proposed in this paper.   

\subsection{Anderson Acceleration for Lloyd's Algorithm}
\label{others}
The key idea of our approach is that a sequence of centroid positions generated by Lloyd's algorithm can be considered as the results of a fixed-point iteration scheme. Since the new centroid positions $\mathbf{C}^{t+1}$ depend on the sample-centroid assignments $\mathbf{P}^{t+1} = \{\rho_i^{t+1}\}$ which in turn depend on the previous centroids $\mathbf{C}^{t}$, we can combine the assignment step and the update step and write $\mathbf{C}^{t+1}$ as a function of $\mathbf{C}^{t}$:
\begin{equation}\label{fixed_point}
	\mathbf{C}^{t+1}_{\mathrm{AU}}=G(\mathbf{C}^{t}).
\end{equation}
Here the subscript AU indicates an iterate using the combined assignment and update step, which we differentiate from the accelerated iterate that will be introduced later. 
If Lloyd's algorithm converges to a local minimum $\mathbf{C}^\ast$, then it must be a fixed point of the mapping $G$, i.e., $\mathbf{C}^\ast = G(\mathbf{C}^\ast)$.
Based on this observation, we can apply Anderson acceleration, a well-established technique for accelerating fixed-point solvers~\citep{Anderson1965,Walker2011}, to improve the convergence of Lloyd's algorithm. Note that Lloyd's algorithm searches for centroid positions $\mathbf{C}$ for which the \emph{residual} $F(\mathbf{C}) = G(\mathbf{C}) - \mathbf{C}$ vanishes. Instead of simply taking $G(\mathbf{C}^{t})$ as the new iterate, Anderson acceleration also makes use of $m$ previous iterates $\mathbf{C}^{t-1}, \ldots, \mathbf{C}^{t-m}$ to determine an accelerated iterate that achieves better reduction of the residual. Let us denote $F^t = F(\mathbf{C}^t), G^t = G(\mathbf{C}^t)$. Then using the iterates $\mathbf{C}^t, \mathbf{C}^{t-1}, \ldots, \mathbf{C}^{t-m}$, we can approximate the high-dimensional graph of $\left(G(\mathbf{C}), F(\mathbf{C})\right)$ with the affine subspace spanned by $(G^t, F^t), (G^{t-1}, F^{t-1}), \ldots, (G^{t-m}, F^{t-m})$. Anderson acceleration searches within this affine space for a point whose residual components are as close to zero as possible, and takes the corresponding $G$ components as the new iterate $\mathbf{C}^{t+1}$. This is equivalent to solving a linear least-squares problem~\citep{Walker2011,Toth2015}
\begin{equation}
\left({\theta}_1^{\ast}, \ldots, {\theta}_m^{\ast}\right)
= \mathop{\arg\min} \left\|F^{t} - \sum_{j=1}^m \theta_j (F^{t-j+1}-F^{t-j}) \right\|^2
\label{eq:lsproblem}
\end{equation} 
and then computing $\mathbf{C}^{t+1}$ as
 \begin{equation}
 \mathbf{C}^{t+1} = G^t + \sum_{j=1}^m \theta_i^\ast (G^{t-j+1}-G^{t-j}).
 \label{eq:AccelIterate}
 \end{equation}
In this way, the accelerated iterate is determined using a more global landscape of the residual, which often results in faster convergence as shown by various authors \citep{Lipnikov2013,Higham2016,Pratapa2016,An2017,Ho2017}. Indeed, it has been shown in~\citep{Fang2009} that Anderson acceleration corresponds to a quasi-Newton method for finding the residual root, which approximates the inverse Jacobian by enforcing secant conditions using $m$ previous iterates.

Although Anderson acceleration works well for many problems, it does not guarantee energy decrease in general, and can stagnate at a wrong solution~\citep{Walker2011,Potra2013}. Recently, \cite{Peng18AA} proposed a modification of Anderson Acceleration that improves stability. For a fixed-point solver that guarantees energy decrease in each iteration, they check the accelerated iterate to see whether it decreases the energy compared with the previous iterate; if not, then they revert to the unaccelerated iterate from the fixed-point solver. This approach guarantees monotonic decrease of the target energy and is shown to be effective in improving robustness of Anderson acceleration. We adopt the same strategy to stabilize the accelerated solver, by checking the decrease of target energy~\eqref{globalEnergy}: if the accelerated iterate does not decrease the energy, then we revert to the iterate using Lloyd's algorithm which guarantees that the new energy value is no larger than the previous iteration. An additional benefit of this strategy is the low computational overhead compared with the unaccelerated solver. For the K-Means clustering problem, the overhead per iteration consists of two parts: (i)~computing the accelerated iterate according to Equations~\eqref{eq:lsproblem} and \eqref{eq:AccelIterate}; (ii)~checking the energy of the accelerated iterate. Part~(i) involves $m$ inner products between $(Kd)$-dimensional vectors as well as solving an $m \times m$ linear system, which is typically a small cost. For Part~(ii), the evaluation of target energy~\eqref{globalEnergy} requires assigning the sample points to the accelerated centroid positions, which can be reused in the next iteration unless we need to revert to the unaccelerated iterate. Therefore, if the accelerated iterate is accepted, then Part~(ii) only involves computing the distance from each sample to its assigned centroid, with a total time complexity of $O(N)$ which is often only a small portion of the computation per iteration~\citep{Peng18AA}. If the accelerated iterate is not accepted, then an additional assignment step will be incurred, but the overhead is still acceptable if we adopt a fast assignment method such as the one from~\citep{hamerly2010making}. In our experiments, the majority of accelerated iterates are accepted (see Tables~\ref{Tab:dynamicM} and \ref{Tab:initialization}), thus the overhead from Part~(ii) is often small.

For the classical Lloyd's algorithm, its convergence is indicated by the same assignment of samples to centroids between two consecutive iterations, because in this case the iterative algorithm can no longer decrease the target energy. For our accelerated algorithm, it is easy to show that the convergence criterion remains the same: since we only accept an accelerated iterate when it decreases the target energy, our algorithm runs until an accelerated iterate is rejected and the fall-back iterate using Lloyd's algorithm results in the same assignment as the previous iteration, which indicates a local minimum of the energy.

\subsection{Dynamic Adjustment of $m$}

\begin{algorithm}[t]
	\KwData{\quad $\mathbf{X}$: sample points; $\mathbf{C}^0$: initial cluster centroids; \\
		\quad $m$: the number of previous iterates used for acceleration; \\
		\quad \assignmentstep{}: assigning samples to clusters according to Eq.~\eqref{eq:AssignmentStep}; \\
		\quad \updatestep{}: updating cluster centroids according to Eq.~\eqref{eq:UpdateStep};\\
		\quad $G$: the mapping combining the assignment and update steps; \\
		\quad $E(\mathbf{P}, \cdot)$: evaluating the K-Means clustering energy~\eqref{globalEnergy} using pre-computed assignment $\mathbf{P}$;\\
		\quad $\overline{m}$: the maximum number of previous iterates used for acceleration.
	}
	\KwResult{A sequence $\{\mathbf{C}^k\}$ converging to a local minimum of $E$.}
	\BlankLine 
	$\mathbf{C}^1 = \mathbf{C}_{\mathrm{AU}}^1 = G(\mathbf{C}^0)$;{~~} $F^0 = \mathbf{C}^1 - \mathbf{C}^0$; {~~} $E^{0} = +\infty$\;
	\For{$t = 1, 2, \ldots$}{
		\tcp{Check for convergence}
		$\mathbf{P}^t = \assignmentstep{}~(\mathbf{X}, \mathbf{C}^t)$\;
		\If{$\mathbf{P}^t == \mathbf{P}^{t-1}$} 
		{
			\Return{$\mathbf{P}^t, \mathbf{C}^t$} \;
		}

		\BlankLine 
		\tcp{Adjust the value of $m$}
		$E^t = E(\mathbf{P}^t, \mathbf{C}^t)$\;
		\uIf{$\frac{E^{t - 1} - E^t}{E^{t - 2} - E^{t - 1}} < \epsilon_1$}
		{
			$m = \max(m - 1, 0)$;
		}
		\ElseIf{$\frac{E^{t - 1} - E^t}{E^{t - 2} - E^{t - 1}} > \epsilon_2$}
		{
			$m = \min(m + 1, \overline{m})$;
		}
		
		\BlankLine 
		\tcp{Make sure $\mathbf{C}^t$ decreases the energy}
		\If{$E^t \geq E^{t-1}$} 
		{
			$\mathbf{C}^t = \mathbf{C}_{\mathrm{AU}}^t$; ~~
			$\mathbf{P}^t = \assignmentstep{}~(\mathbf{X}, \mathbf{C}_{\mathrm{AU}}^t)$; ~~
			$E^t = E(\mathbf{P}^t, \mathbf{C}^t)$\;
		}
		
		\BlankLine 
		\tcp{Anderson acceleration}
		$\mathbf{C}_{\mathrm{AU}}^t = \updatestep{}~(\mathbf{X}, \mathbf{P}^t)$\; 
		$G^t = \mathbf{C}_{\mathrm{AU}}^t$; {~~}
		$F^t = G^t - \mathbf{C}^{t}$; {~~}
		$m_t = \min(m, t)$\;
		$(\theta_1^\ast, \ldots, \theta_{m_t}^\ast) = \argmin \|F^t - \sum_{j=1}^{m_t} \theta_j (F^{t-j+1}-F^{t-j})  \|^2$\;
		$\mathbf{C}^{t+1}  = G^t - \sum_{j=1}^{m_t} \theta_j^\ast (G^{t-j+1}-G^{t-j})$;
	}
	\caption{Anderson acceleration for the K-Means algorithm.}
	\label{alg:AA-Kmeans}
\end{algorithm}

By default, the number of previous iterates $m$ used in Anderson acceleration is fixed during the whole solving process, and can influence the convergence rate~\citep{Higham2016,Peng18AA}. With a larger value of $m$, more information is utilized for approximating the inverse Jacobian. On the other hand, a larger $m$ increases the computational cost, and may overfit iterates that are far away. In general, the optimal choice of $m$ is problem-dependent and is difficult to determine beforehand. To achieve more robust speedup, we propose an adaptive strategy to dynamically adjust the value of $m$ according to the decrease of energy. The key idea is to compare the energy decrease from the current iterate with the decrease from the last iterate. In particular:   
\begin{itemize}[leftmargin=*]
\item If the current iterate increases the energy, or the energy decrease in the current iteration is small compared to the previous iteration, then $m$ should be become smaller. 
\item If the energy decrease in the current iteration is large enough compared to the previous iteration, then $m$ should be larger.
\item Otherwise, $m$ is not altered.
\end{itemize}
This strategy is similar to the trust region method that enlarges or shrinks the trust region according to the effectiveness of the current step~\citep{NoceWrig06}. The full acceleration algorithm for K-Means clustering is shown in Algorithm~\ref{alg:AA-Kmeans}, and lines 4-8 are the adjustment steps for the $m$ value. Our experiments show that dynamic adjustment of $m$ achieves better performance than a fixed $m$ value for many problem instances (see Section~\ref{sec:MSetting}).

%% file: results.tex
\section{Results}

We evaluate the performance of our algorithm and compare it with other K-means clustering algorithms in a series of experiments.
All algorithms are implemented in C++ and parallelized on the CPU using OpenMP. The \assignmentstep{} in Algorithm~\ref{alg:AA-Kmeans} is implemented based on Hamerly's method~\citep{hamerly2010making}. Although the it can be further optimized using more recent methods such as those from~\citep{ding2015yinyang} and~\citep{newling2016fast}, this will not affect the reduction ratio in the number of iterations.

We conduct the experiments on 20 datasets in a variety of sizes. They consist of 19 real-world datasets from the UCI machine learning repository~\citep{Dua:2017}, and the synthetic Birch dataset~\citep{Birchsets} that contains clusters in regular grid structure\footnote{https://cs.joensuu.fi/sipu/datasets/}. Table~\ref{Tab:datasets} shows the size and dimension of each data set ($N$ for the number of samples, $d$ for the dimension). All algorithms are started with the same initial centroids and iterated until convergence, and are tested on a desktop PC with a quad-core Intel CPU i7 and 4GB RAM. For Algorithm~\ref{alg:AA-Kmeans}, in all tests we set $\varepsilon_{1} = 0.02$ and $\varepsilon_{2} = 0.5$, $\overline{m} = 30$, and initialize $m$ to 2 by default.

\subsection{Settings of $m$}
\label{sec:MSetting}

Table~\ref{Tab:dynamicM} compares the performance of our method using fixed and dynamic $m$ values, showing the computational time and the number of iterations required by each strategy. We can observe that in the majority of cases, dynamic adjustment of $m$ reduces the computational time as well as the number of iterations compared with a fixed $m$ strategy. 
The total computational time is reduced by more than 20 percent for most datasets, and even better for some datasets such as $Slicelocalization (\# 2)$, $Letterrecognition (\# 4)$ and $MiniBoone (\# 10)$. Although we use the same values of $\varepsilon_1$ and $\varepsilon_2$ for all datasets, the dynamic adjustment strategy performs consistently better than a fixed $m$ strategy.

\begin{table}[!t]
        \caption{The 20 datasets used in our experiments.}
        \label{Tab:datasets}
        \centering
        \begin{small}
        \begin{tabular}{ | c  c  c  c  ||  c  c  c  c |}
        \Xhline{1pt}
        No. & Name	 & $N$	& $d$ & No.	& Name & $N$	& $d$ \\
        \hline       
       	1	&  UCIHARDATAXtrain & 7352 & 561 & 11 & Colorment & 68040 & 9
        \\
        2	&  Slicelocalization & 53500 & 385 & 12 & Conflongdemo & 164860 & 3
        \\
        3 & RelationNetwork & 53413 & 22 & 13 & Birch & 100000 & 2
        \\
        4 & Letterrecognition & 20000 & 16 & 14 & Shuttle & 43500 & 9
        \\
        5 & HTRU2 & 17898 & 8 & 15 & Covtype & 581012 & 55
        \\
        6 & Household & 2049280 & 6 & 16 & SkinNonSkin & 245057 & 4
        \\
        7 & FrogsMFCCs & 7195 & 21 & 17 & Finalgeneral& 10104 & 72
        \\
        8 & Eb & 45781 & 2 & 18 & ColorHistogram & 68040 & 32
        \\
        9 & AllUsers & 78095 & 8 & 19 & USCensus1990 & 2458285 & 69
        \\
        10 & MiniBoone & 130064 & 50 & 20 & Kddcup99 & 4898431 & 37
        \\\Xhline{1pt}
        
        \end{tabular}
        \end{small}
\end{table}

\begin{table}[!t]
        \caption{Performance comparison between strategies of fixed and dynamic $m$ value for our method. For dynamic strategies, the initial $m$ value is shown in the header. For each strategy, the number $a/b$ in the column \#Iter means that the solver takes $b$ iterations, out of which $a$ iterations accepts the accelerated iterate. Between each pair of strategies with the same $m$, the shortest computational time for each dataset is highlighted in bold font.}
        \label{Tab:dynamicM}
        \setlength{\tabcolsep}{1.6pt}
        \centering
        \begin{small}
        \begin{tabular}{ | c ||  c  c  c | c  c  c  || c  c  c | c  c  c | }
        \Xhline{1pt}
        \multicolumn{1}{ | c ||}{\multirow{2}{*}{Dataset}}
        &\multicolumn{3}{ c |}{Fixed $m=2$}	
        &\multicolumn{3}{ c ||}{Dynamic $m=2$}	
        &\multicolumn{3}{ c |}{Fixed $m=5$}	
        &\multicolumn{3}{ c | }{Dynamic $m=5$}\\\cline{2-13}
        \multicolumn{1}{ | c ||}{} &
        \#Iter		&Time  (s) 	&MSE	&
        \#Iter		&Time  (s)	 &MSE	&
        \#Iter		&Time  (s) 	&MSE	&
        \#Iter		&Time  (s)	 &MSE\\\hline
        
        1 & 22 / 27 & \textbf{0.22} & 15.08 &
        27 / 31 & 0.25 & 15.08 &
        36 / 45 & 0.46 & 15.02 &
        27 / 35 & \textbf{0.36} & 15.03 \\
        
        2 & 27 / 30 & 1.36 & 15.16 &
        18 / 21 & \textbf{0.98} & 15.16 &
        32 / 38 & 1.97 & 15.16 &
        26 / 29 & \textbf{1.40} & 15.16 \\
        
        3 & 39 / 46 & 0.48 & 2.61 &
        31 / 38 & \textbf{0.43} & 2.61 &
        12 / 13 & \textbf{0.14} & 2.63 &
        12 / 13 & 0.15 & 2.63 \\
        
        4 & 50 / 61 & 0.30 & 2.89 &
        17 / 22 & \textbf{0.11} & 2.89 &
        35 / 38 & 0.20 & 2.90 &
        19 / 21 & \textbf{0.11} & 2.90 \\
        
        5 & 38 / 49 & 0.15 & 1.24 &
        23 / 26 & \textbf{0.07} & 1.24 &
        38 / 54 & 0.21 & 1.24 &
        34 / 43 & \textbf{0.15} & 1.24 \\
        
        6 & 30 / 35 & 11.19 & 1.17 &
        26 / 29 & \textbf{9.21} & 1.17 &
        18 / 22 & 8.97 & 1.16 &
        16 / 18 & \textbf{6.89} & 1.16 \\
        
        7 & 20 / 21 & 0.03 & 2.57 &
        16 / 18 & \textbf{0.02} & 2.57 &
        12 / 13 & 0.02 & 2.59 &
        10 / 11 & \textbf{0.01} & 2.59 \\
        
        8 & 30 / 35 & \textbf{0.12} & 0.45 &
        35 / 46 & 0.20 & 0.45 &
        21 / 23 & 0.08 & 0.45 &
        14 / 16 & \textbf{0.07} & 0.45 \\
        
        9 & 41 / 51 & 0.88 & 1.93 &
        34 / 43 & \textbf{0.77} & 1.93 &
        38 / 55 & 1.21 & 1.93 &
        36 / 53 & \textbf{1.21} & 1.93 \\
        
        10 & 39 / 41 & 1.30 & 1.70 &
        26 / 29 & \textbf{0.96} & 1.70 &
        27 / 34 & 1.23 & 1.70 &
        19 / 22 & \textbf{0.76} & 1.70 \\
        
        11 & 99 / 119 & 1.71 & 2.06 &
        71 / 83 & \textbf{1.16} & 2.06 &
        139 / 179 & 2.74 & 2.06 &
        91 / 115 & \textbf{1.68} & 2.06 \\
        
        12 & 16 / 18 & \textbf{0.39} & 0.83 &
        19 / 21 & 0.45 & 0.83 &
        19 / 20 & 0.45 & 0.83 &
        18 / 19 & \textbf{0.41} & 0.83 \\
        
        13 & 14 / 21 & \textbf{0.19} & 0.42 &
        17 / 24 & 0.22 & 0.42 &
        9 / 11 & 0.08 & 0.41 &
        10 / 11 & \textbf{0.08} & 0.41 \\
        
        14 & 6 / 9 & 0.07 & 1.63 &
        6 / 9 & \textbf{0.06} & 1.63 &
        7 / 9 & 0.06 & 1.68 &
        7 / 9 & \textbf{0.06} & 1.68 \\
        
        15 & 22 / 25 & \textbf{3.86} & 6.47 &
        20 / 24 & 3.89 & 6.47 &
        21 / 26 & 4.39 & 6.47 &
        17 / 22 & \textbf{3.82} & 6.47 \\
        
        16 & 9 / 12 & 0.30 & 0.55 &
        8 / 11 & \textbf{0.28} & 0.55 &
        12 / 16 & 0.39 & 0.55 &
        12 / 16 & \textbf{0.39} & 0.55 \\
        
        17 & 46 / 47 & 0.13 & 7.63 &
        29 / 32 & \textbf{0.10} & 7.63 &
        43 / 53 & 0.22 & 7.63 &
        18 / 19 & \textbf{0.07} & 7.63 \\
        
        18 & 19 / 22 & \textbf{0.37} & 4.53 &
        21 / 26 & 0.47 & 4.53 &
        29 / 35 & 0.72 & 4.53 &
        16 / 19 & \textbf{0.38} & 4.53 \\
        
        19 & 14 / 15 & \textbf{14.37} & 6.02 &
        15 / 17 & 16.91 & 6.02 &
        14 / 20 & 22.50 & 6.02 &
        15 / 21 & \textbf{21.09} & 6.02 \\
        
        20 & 8 / 10 & 6.11 & 3.91 &
        8 / 10 & \textbf{6.04} & 3.91 &
        8 / 12 & 7.71 & 3.78 &
        9 / 12 & \textbf{7.33} & 3.78 \\\Xhline{1pt}     
        \end{tabular}
        \end{small}
\end{table}

\subsection{Comparison with Lloyd's Algorithm}

In Table~\ref{Tab:initialization}, we compare our method with Lloyd's algorithm using the assignment method from \citet{hamerly2010making}. We show the number of iterations, the total computational time, and the final mean squared error (MSE) between the samples and their corresponding centroids. To demonstrate the robustness of our method to the initial centroid positions, we test four different initialization techniques: K-Means++~\citep{arthur2007k}, afk-$mc^{2}$~\citep{bachem2016fast}, bf~\citep{bradley1998refining}, and CLARANS~\citep{newling2017k}. The initial centroids are generated by the code accompanying the paper~\citep{newling2017k}.

For all 20 datasets, we test the methods by setting the number of clusters $K=10$ and starting from each of the four sets of initial centroids, resulting in 80 test cases. Table~\ref{Tab:initialization} shows that our method requires less computational time than Lloyd's algorithm in the majority of cases. In particular, our method reduces the computational time by more than 25\% in 42 cases, and more than 50\% in 15 cases. With the initialization techniques K-Means++, afk-$mc^{2}$, bf, and CLARANS, our method outperforms Lloyd's algorithm in 15, 20, 18, and 19 datasets out of 20, respectively.

To demonstrate the robustness of our method to the number of clusters $K$, we also test each dataset using different values of $K$. The last three columns of Table~\ref{Tab:initialization} shows the performance of both methods initialized with CLARANS and for cluster numbers $K=10$, $K=100$, and $K=1000$, respectively. We can observe that our method consistently outperforms Lloyds' algorithm for different $K$ values.

\begin{table}[!htbp]
        \caption{Comparison between our method and \citep{hamerly2010making}, for different initialization strategies and different numbers of clusters. In each cell, the performance of our method is show on the right.} 
        \label{Tab:initialization} 
        \setlength{\tabcolsep}{1.3pt}
        \centering
        \begin{small}
     	\begin{tabular}{  | c | c | c  c | c  c | c  c | c  c | c  c | c c | }
        \Xhline{1pt}
        \multicolumn{2}{ | c | }{\multirow{2}{*}{Dataset}}
        &\multicolumn{2}{ c |}{K-Means++}	
        & \multicolumn{2}{ c |}{afk-$mc^{2}$}
        &\multicolumn{2}{ c |}{bf}
        & \multicolumn{6}{ c |}{CLARANS} \\ \cline{9-14}
        \multicolumn{2}{ | c | }{}
        &\multicolumn{2}{ c |}{$(K=10)$}	
        & \multicolumn{2}{ c |}{$(K=10)$}
        &\multicolumn{2}{ c |}{$(K=10)$}
        & \multicolumn{2}{ c |}{$K=10$}
        & \multicolumn{2}{ c |}{$K=100$}
        & \multicolumn{2}{ c | }{$K=1000$} \\ \hline
\multirow{3}*{1}
& \#Iter       & 26  & \textbf{22} / \textbf{25} & 39  & \textbf{21} / \textbf{23} & 23  & \textbf{12} / \textbf{14} & \textbf{30}  & 27 / 31 & 53  & \textbf{26} / \textbf{27} & 14  & \textbf{10} / \textbf{11} \\
& Time (s)   & 0.31  & \textbf{0.30} & 0.43  & \textbf{0.29} & 0.33  & \textbf{0.16} & 0.35  & \textbf{0.25} & 5.74  & \textbf{2.97} & 19.38  & \textbf{14.41} \\
& MSE        & 15.25  & 15.25 & 15.03  & 15.03 & 15.97  & 15.97 & 15.08  & 15.08 & 12.40  & 12.41 & 9.63  & 9.63 \\\hline
\multirow{3}*{2}
& \#Iter       & 87  & \textbf{61} / \textbf{79} & 27  & \textbf{22} / \textbf{24} & 15  & \textbf{10} / \textbf{11} & 42  & \textbf{18} / \textbf{21} & \textbf{49}  & 49 / 55 & 40  & \textbf{31} / \textbf{33} \\
& Time (s)   & \textbf{5.32}  & 5.71 & 1.55  & \textbf{1.37} & 0.83  & \textbf{0.61} & 2.40  & \textbf{0.98} & \textbf{28.14}  & 34.28 & 167.11  & \textbf{143.06} \\
& MSE        & 15.26  & 15.26 & 15.27  & 15.27 & 16.78  & 16.78 & 15.16  & 15.16 & 12.55  & 12.55 & 8.72  & 8.72 \\\hline
\multirow{3}*{3}
& \#Iter       & 31  & \textbf{23} / \textbf{27} & 33  & \textbf{20} / \textbf{24} & 28  & \textbf{16} / \textbf{18} & 54  & \textbf{31} / \textbf{38} & 41  & \textbf{26} / \textbf{28} & 20  & \textbf{16} / \textbf{17} \\
& Time (s)   & 0.49  & \textbf{0.46} & 0.42  & \textbf{0.35} & 0.41  & \textbf{0.22} & 0.68  & \textbf{0.43} & 3.54  & \textbf{2.50} & 13.58  & \textbf{11.44} \\
& MSE        & 2.68  & 2.68 & 2.74  & 2.74 & 2.77  & 2.77 & 2.61  & 2.61 & 1.47  & 1.47 & 0.51  & 0.51 \\\hline
\multirow{3}*{4}
& \#Iter       & \textbf{51}  & 51 / 61 & 57  & \textbf{31} / \textbf{35} & 75  & \textbf{42} / \textbf{50} & 66  & \textbf{17} / \textbf{22} & 98  & \textbf{43} / \textbf{49} & 21  & \textbf{15} / \textbf{16} \\
& Time (s)   & \textbf{0.32}  & 0.42 & 0.34  & \textbf{0.21} & 0.39  & \textbf{0.29} & 0.39  & \textbf{0.11} & 4.57  & \textbf{2.51} & 9.69  & \textbf{7.42} \\
& MSE        & 2.90  & 2.90 & 2.92  & 2.92 & 2.90  & 2.90 & 2.89  & 2.89 & 1.88  & 1.88 & 1.13  & 1.13 \\\hline
\multirow{3}*{5}
& \#Iter       & 137  & \textbf{78} / \textbf{104} & 110  & \textbf{38} / \textbf{52} & 127  & \textbf{63} / \textbf{82} & 51  & \textbf{23} / \textbf{26} & 65  & \textbf{26} / \textbf{31} & 28  & \textbf{25} / \textbf{28} \\
& Time (s)   & 0.50  & \textbf{0.46} & 0.38  & \textbf{0.25} & 0.40  & \textbf{0.32} & 0.20  & \textbf{0.07} & 2.13  & \textbf{1.15} & \textbf{10.72}  & 11.46 \\
& MSE        & 1.24  & 1.24 & 1.24  & 1.24 & 1.24  & 1.24 & 1.24  & 1.24 & 0.62  & 0.62 & 0.31  & 0.31 \\\hline
\multirow{3}*{6}
& \#Iter       & 209  & \textbf{53} / \textbf{89} & 241  & \textbf{38} / \textbf{55} & 67  & \textbf{29} / \textbf{33} & 61  & \textbf{26} / \textbf{29} & 642  & \textbf{282} / \textbf{377} & \textbf{403}  & 549 / 610 \\
& Time (s)   & 87.86  & \textbf{54.79} & 96.47  & \textbf{30.81} & 27.67  & \textbf{14.78} & 22.48  & \textbf{9.21} & 2043.86  & \textbf{1546.83} & \textbf{10350.36}  & 17077.98 \\
& MSE        & 1.18  & 1.18 & 1.20  & 1.20 & 1.33  & 1.33 & 1.17  & 1.17 & 0.63  & 0.63 & 0.29  & 0.29 \\\hline
\multirow{3}*{7}
& \#Iter       & \textbf{15}  & 14 / 17 & 23  & \textbf{13} / \textbf{15} & 12  & \textbf{9} / \textbf{10} & 22  & \textbf{16} / \textbf{18} & 70  & \textbf{35} / \textbf{46} & 17  & \textbf{12} / \textbf{13} \\
& Time (s)   & \textbf{0.02}  & 0.03 & 0.04  & \textbf{0.03} & 0.02  & \textbf{0.01} & 0.04  & \textbf{0.02} & 1.42  & \textbf{1.16} & 4.47  & \textbf{3.42} \\
& MSE        & 2.63  & 2.63 & 2.84  & 2.84 & 2.91  & 2.91 & 2.57  & 2.57 & 1.62  & 1.62 & 0.98  & 0.98 \\\hline
\multirow{3}*{8}
& \#Iter       & 97  & \textbf{51} / \textbf{67} & 87  & \textbf{24} / \textbf{28} & 83  & \textbf{20} / \textbf{30} & 82  & \textbf{35} / \textbf{46} & 175  & \textbf{82} / \textbf{89} & \textbf{33}  & 32 / 35 \\
& Time (s)   & 0.48  & \textbf{0.40} & 0.44  & \textbf{0.16} & 0.36  & \textbf{0.18} & 0.40  & \textbf{0.20} & 5.86  & \textbf{3.20} & \textbf{13.53}  & 14.43 \\
& MSE        & 0.45  & 0.45 & 0.45  & 0.45 & 0.45  & 0.45 & 0.45  & 0.45 & 0.14  & 0.14 & 0.04  & 0.04 \\\hline
\multirow{3}*{9}
& \#Iter       & 106  & \textbf{57} / \textbf{74} & 101  & \textbf{35} / \textbf{45} & 57  & \textbf{24} / \textbf{28} & 65  & \textbf{34} / \textbf{43} & 83  & \textbf{31} / \textbf{37} & 35  & \textbf{30} / \textbf{32} \\
& Time (s)   & 2.25  & \textbf{1.93} & 2.08  & \textbf{1.09} & 1.22  & \textbf{0.68} & 1.36  & \textbf{0.77} & 14.51  & \textbf{7.34} & 43.43  & \textbf{39.93} \\
& MSE        & 1.94  & 1.94 & 1.93  & 1.93 & 2.19  & 2.19 & 1.93  & 1.93 & 1.27  & 1.27 & 0.73  & 0.73 \\\hline
\multirow{3}*{10}
& \#Iter       & 142  & \textbf{91} / \textbf{125} & 85  & \textbf{41} / \textbf{53} & 37  & \textbf{14} / \textbf{16} & 57  & \textbf{26} / \textbf{29} & 361  & \textbf{65} / \textbf{80} & 129  & \textbf{48} / \textbf{53} \\
& Time (s)   & \textbf{6.40}  & 7.11 & 3.36  & \textbf{2.62} & 1.12  & \textbf{0.50} & 2.33  & \textbf{0.96} & 150.54  & \textbf{41.07} & 430.11  & \textbf{193.56} \\
& MSE        & 1.41  & 1.41 & 1.70  & 1.70 & 1.85  & 1.85 & 1.70  & 1.70 & 0.87  & 0.87 & 0.60  & 0.60 \\\hline
\multirow{3}*{11}
& \#Iter       & 106  & \textbf{72} / \textbf{88} & 169  & \textbf{69} / \textbf{93} & 89  & \textbf{26} / \textbf{32} & 142  & \textbf{71} / \textbf{83} & 316  & \textbf{81} / \textbf{88} & 95  & \textbf{59} / \textbf{61} \\
& Time (s)   & 1.52  & \textbf{1.49} & 2.09  & \textbf{1.45} & 1.32  & \textbf{0.57} & 1.83  & \textbf{1.16} & 42.87  & \textbf{13.10} & 128.37  & \textbf{83.78} \\
& MSE        & 2.05  & 2.05 & 2.06  & 2.06 & 2.19  & 2.19 & 2.06  & 2.06 & 1.38  & 1.38 & 0.91  & 0.91 \\\hline
\multirow{3}*{12}
& \#Iter       & 73  & \textbf{28} / \textbf{44} & 38  & \textbf{24} / \textbf{28} & 48  & \textbf{28} / \textbf{35} & 36  & \textbf{19} / \textbf{21} & 190  & \textbf{81} / \textbf{92} & 201  & \textbf{102} / \textbf{110} \\
& Time (s)   & 2.06  & \textbf{1.79} & 1.06  & \textbf{0.94} & 1.40  & \textbf{1.26} & 0.94  & \textbf{0.45} & 41.45  & \textbf{22.73} & 353.33  & \textbf{215.20} \\
& MSE        & 0.83  & 0.83 & 0.85  & 0.85 & 0.91  & 0.91 & 0.83  & 0.83 & 0.40  & 0.40 & 0.18  & 0.18 \\\hline
\multirow{3}*{13}
& \#Iter       & 25  & \textbf{13} / \textbf{18} & 27  & \textbf{16} / \textbf{18} & 27  & \textbf{11} / \textbf{14} & 39  & \textbf{17} / \textbf{24} & 101  & \textbf{44} / \textbf{57} & 61  & \textbf{33} / \textbf{35} \\
& Time (s)   & 0.34  & \textbf{0.31} & 0.39  & \textbf{0.30} & 0.30  & \textbf{0.17} & 0.39  & \textbf{0.22} & 7.06  & \textbf{4.74} & 56.28  & \textbf{32.52} \\
& MSE        & 0.45  & 0.45 & 0.46  & 0.46 & 0.42  & 0.42 & 0.42  & 0.42 & 0.09  & 0.09 & 0.03  & 0.03 \\\hline
\multirow{3}*{14}
& \#Iter       & 30  & \textbf{19} / \textbf{22} & 39  & \textbf{20} / \textbf{28} & 10  & \textbf{7} / \textbf{9} & 19  & \textbf{6} / \textbf{9} & 25  & \textbf{7} / \textbf{9} & 25  & \textbf{16} / \textbf{17} \\
& Time (s)   & 0.16  & \textbf{0.14} & 0.25  & \textbf{0.25} & \textbf{0.07}  & 0.08 & 0.14  & \textbf{0.06} & 1.42  & \textbf{0.65} & 16.23  & \textbf{11.32} \\
& MSE        & 1.77  & 1.77 & 1.71  & 1.71 & 1.69  & 1.69 & 1.63  & 1.63 & 0.35  & 0.35 & 0.14  & 0.14 \\\hline
\multirow{3}*{15}
& \#Iter       & 48  & \textbf{20} / \textbf{25} & 113  & \textbf{60} / \textbf{80} & 33  & \textbf{14} / \textbf{19} & 30  & \textbf{20} / \textbf{24} & 62  & \textbf{35} / \textbf{39} & 226  & \textbf{150} / \textbf{168} \\
& Time (s)   & 10.36  & \textbf{6.23} & 21.38  & \textbf{19.96} & 6.81  & \textbf{4.75} & 5.91  & \textbf{3.89} & 100.54  & \textbf{67.53} & 3087.60  & \textbf{2541.44} \\
& MSE        & 6.50  & 6.50 & 6.51  & 6.51 & 6.48  & 6.48 & 6.47  & 6.47 & 2.69  & 2.69 & 1.30  & 1.30 \\\hline
\multirow{3}*{16}
& \#Iter       & 12  & \textbf{9} / \textbf{10} & 47  & \textbf{19} / \textbf{29} & 34  & \textbf{15} / \textbf{22} & 20  & \textbf{8} / \textbf{11} & 42  & \textbf{20} / \textbf{24} & \textbf{25}  & 35 / 39 \\
& Time (s)   & 0.37  & \textbf{0.28} & 1.35  & \textbf{1.15} & 0.77  & \textbf{0.72} & 0.48  & \textbf{0.28} & 7.84  & \textbf{5.19} & \textbf{74.01}  & 110.09 \\
& MSE        & 0.60  & 0.60 & 0.63  & 0.63 & 0.72  & 0.72 & 0.55  & 0.55 & 0.15  & 0.15 & 0.05  & 0.05 \\\hline
\multirow{3}*{17}
& \#Iter       & \textbf{31}  & 28 / 36 & 49  & \textbf{25} / \textbf{28} & 3  & \textbf{2} / \textbf{3} & 58  & \textbf{29} / \textbf{32} & 35  & \textbf{18} / \textbf{22} & 15  & \textbf{12} / \textbf{13} \\
& Time (s)   & \textbf{0.15}  & 0.18 & 0.21  & \textbf{0.13} & 0.01  & \textbf{0.01} & 0.25  & \textbf{0.10} & 1.37  & \textbf{1.03} & 6.81  & \textbf{5.60} \\
& MSE        & 7.62  & 7.62 & 7.74  & 7.74 & 7.78  & 7.78 & 7.63  & 7.63 & 6.12  & 6.12 & 4.70  & 4.69 \\\hline
\multirow{3}*{18}
& \#Iter       & 44  & \textbf{28} / \textbf{33} & 75  & \textbf{31} / \textbf{46} & 55  & \textbf{25} / \textbf{28} & 43  & \textbf{21} / \textbf{26} & 163  & \textbf{46} / \textbf{51} & \textbf{77}  & 83 / 85 \\
& Time (s)   & 1.03  & \textbf{0.77} & 1.51  & \textbf{1.20} & 1.15  & \textbf{0.69} & 0.91  & \textbf{0.47} & 31.06  & \textbf{10.33} & \textbf{120.59}  & 134.91 \\
& MSE        & 4.53  & 4.53 & 4.55  & 4.55 & 5.23  & 5.23 & 4.53  & 4.53 & 2.61  & 2.61 & 1.78  & 1.78 \\\hline
\multirow{3}*{19}
& \#Iter       & 69  & \textbf{34} / \textbf{60} & 57  & \textbf{23} / \textbf{34} & 15  & \textbf{11} / \textbf{13} & 17  & \textbf{15} / \textbf{17} & 60  & \textbf{32} / \textbf{32} & 1  & \textbf{46} / \textbf{1} \\
& Time (s)   & 255.37  & \textbf{72.35} & 52.41  & \textbf{41.87} & 13.22  & \textbf{12.22} & \textbf{16.66}  & 16.91 & 543.64  & \textbf{288.18} & 754.89  & \textbf{630.99} \\
& MSE        & 6.17  & 6.17 & 6.11  & 6.11 & 6.66  & 6.66 & 6.02  & 6.02 & 4.18  & 4.18 & 0.01  & 0.00 \\\hline
\multirow{3}*{20}
& \#Iter       & 13  & \textbf{11} / \textbf{12} & 20  & \textbf{13} / \textbf{15} & \textbf{8}  & 9 / 10 & 14  & \textbf{8} / \textbf{10} & 83  & \textbf{38} / \textbf{53} & 163  & \textbf{58} / \textbf{72} \\
& Time (s)   & 11.09  & \textbf{11.09} & 14.83  & \textbf{13.37} & \textbf{5.84}  & 6.80 & 7.72  & \textbf{6.04} & 220.72  & \textbf{191.36} & 10408.44  & \textbf{7212.02} \\
& MSE        & 3.92  & 3.92 & 4.07  & 4.07 & 3.89  & 3.89 & 3.91  & 3.91 & 1.86  & 1.86 & 0.77  & 0.77\\\Xhline{1pt}
        
        \end{tabular}
        \end{small}
\end{table}

%% file: conclusion.tex
\section{Conclusion}
\label{conclusion}

In this paper, we apply a modified Anderson acceleration scheme to improve the performance of Lloyd's algorithm. We also propose a dynamic adjustment strategy of the parameter $m$ to improve the robustness of Anderson acceleration.
Experimental results show that our method can improve the convergence rate of Lloyd's algorithm regardless of the dimension and size of the data sets, the number of clusters, and initial values. The proposed method can be combined with existing acceleration schemes that reduce the assignment cost and improve the initialization. 

Although our algorithm performs consistently better than classical Lloyd's algorithm, its acceleration ratio varies between different problem instances. One interesting future work would be to investigate characteristics of samples that influence the effectiveness of our approach. In addition, many optimization algorithms used in machine learning have similar forms as Lloyd's algorithm. Therefore, another interesting direction for future research is to extend the proposed algorithm to other problems.